\pdfoutput=1

\documentclass[11pt]{article}

\usepackage[preprint]{acl}

\usepackage{times}
\usepackage{latexsym}

\usepackage[T1]{fontenc}

\usepackage[utf8]{inputenc}

\usepackage{microtype}

\usepackage{inconsolata}

\usepackage{graphicx}

\usepackage{amsmath}
\usepackage{amsfonts}
\usepackage{bbm}
\usepackage{caption}

\newcommand{\daniel}[1]{\textcolor{cyan}{[df: #1]}}
\newcommand{\andre}[1]{\textcolor{orange}{[ah: #1]}}

\title{Rewarding the Unlikely: 
Lifting GRPO \\ Beyond Distribution Sharpening}

\author{Andre He \\
  Carnegie Mellon University \\\And
  Daniel Fried \\
  Carnegie Mellon University \\\And
  Sean Welleck \\
  Carnegie Mellon University \\}

\begin{document}
\maketitle
\begin{abstract}
Reinforcement learning is emerging as a primary driver for improving language model reasoning capabilities. A fundamental question is whether current reinforcement learning algorithms---such as Group Relative Policy Optimization (GRPO), the \textit{de facto} standard algorithm used to improve language model reasoning---merely sharpen the base model's distribution around problems it can already solve.
We investigate this question in the context of formal theorem proving, which has access to a perfect verifier.
We identify a degenerate \textit{rank bias} in  GRPO in which highly probable trajectories are reinforced and rare ones are neglected.
This results in distribution sharpening: the model can solve some problems with fewer samples, but underperforms simply sampling more solutions from the original model.
To overcome GRPO's rank bias we introduce \textit{unlikeliness reward}, a simple method for explicitly up-weighting rare but correct solutions.
We show that unlikeliness reward mitigates rank bias and improves pass@$N$ across a large range of $N$ in both synthetic and real theorem proving settings.
We also uncover an unexpected link between rank bias and a seemingly mundane hyperparameter---the number of updates per batch---that leads to a second, complementary mitigation.
We combine our insights into a revised GRPO  training recipe for formal theorem proving, yielding an open pipeline that achieves competitive performance to DeepSeek-Prover-V1.5-RL on the miniF2F-test benchmark.
We release our implementation at \url{https://github.com/AndreHe02/rewarding-unlikely-release}.
\end{abstract}

\begin{figure}[t]
    \centering
    \includegraphics[width=\columnwidth]{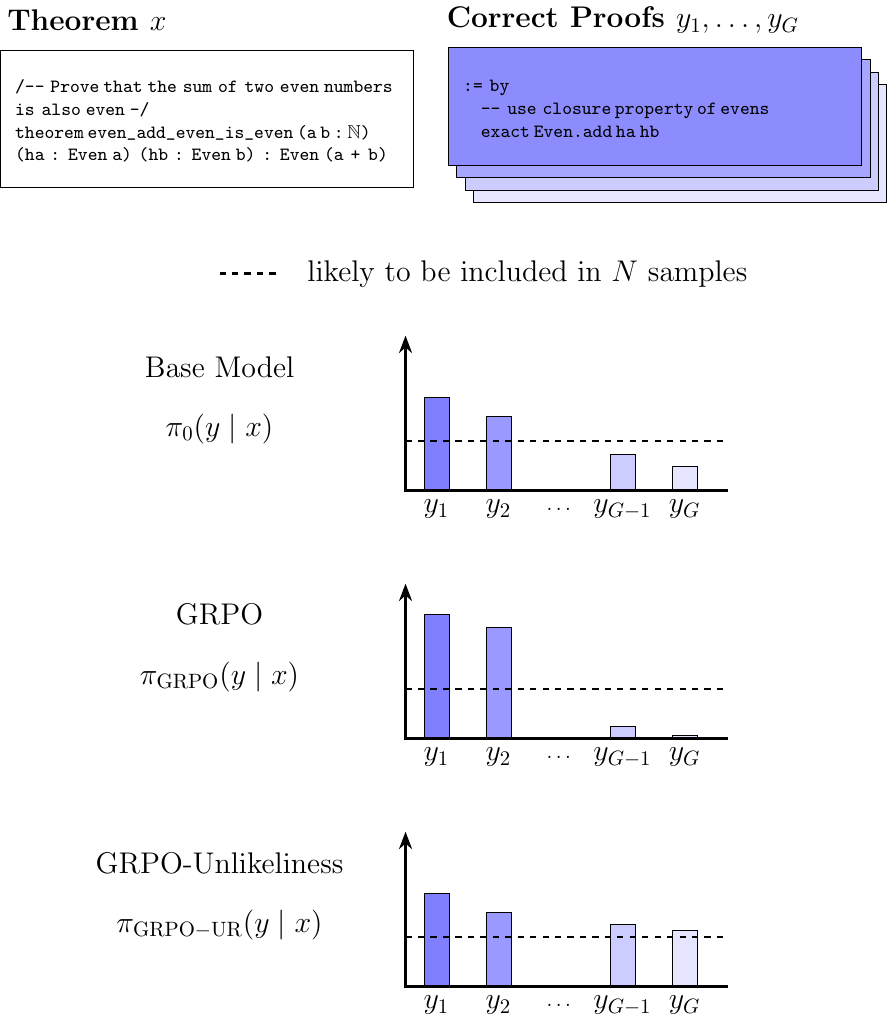}
    \caption{We identify a \textit{rank bias} in GRPO in which model updates only reinforce already probable solutions and fail to surface new ones. This sharpens the distribution and impairs pass@$N$ performance for large N. Our 
    \textit{unlikeliness reward} addresses rank bias by 
    explicitly encouraging uplifting low-probability correct solutions.}
    \label{fig:mainfigure}
    \vspace{-1.0em}
\end{figure}

\section{Introduction}

Reinforcement learning (RL) has recently emerged as a powerful framework for enhancing the reasoning capabilities of large language models (LLMs). In domains such as mathematics and code generation, RL has been applied at scale to elicit complex reasoning behaviors using only problem instances and their corresponding outcome rewards \citep{deepseekai2025deepseekr1incentivizingreasoningcapability, yu2025dapoopensourcellmreinforcement}. 

Formal theorem proving is a particularly attractive domain for studying LLM reasoning. Formal systems such as Lean and Isabelle \citep{lean, isabella} can verify mathematical proofs step-by-step, ensuring that models are only rewarded for fully correct solutions. Since verification is fully automated and immune to spurious solutions, formal mathematics serves as an ideal testbed for reinforcement learning algorithms.

An important open challenge is designing reinforcement learning algorithms that do more than ``sharpen the distribution''---that is, we want the RL-trained model to solve problems that cannot be solved by simply sampling more from the original model. Consistent with the findings of \citet{yue2025doesreinforcementlearningreally}, our initial experiments identify this as a key limitation of existing RL recipes based on Group Relative Policy Optimization (GRPO)~\citep{shao2024deepseekmathpushinglimitsmathematical}, the de facto standard algorithm for improving LLM reasoning. While GRPO improves single-sample accuracy, it often fails to improve and can even impair pass@$N$ metrics at larger $N$ in our theorem proving setting (\autoref{fig:initresults}). This is a significant limitation in domains with a perfect verifier, such as formal mathematics, since these domains naturally lend themselves to 
sampling and verifying many candidates at test time.

We argue that improving pass@$N$ performance requires specifically increasing the probability of \textit{low probability correct responses} under the model. We construct a toy model to demonstrate this phenomenon, and reveal empirically that GRPO suffers from \textit{rank bias}: a tendency to reinforce already high-likelihood responses while neglecting the long tail of rare but correct ones. This reduces sample diversity and degrades multi-sample performance over time. To address this, we introduce \textbf{Unlikeliness Reward}, which up-weights correct outputs that are less likely than others. 
Doing so dramatically changes how GRPO learns from less likely trajectories, translating to more output diversity and higher pass@$N$ across a range of $N$ values.

Furthermore, we uncover an unexpected link between GRPO's distribution sharpening and a seemingly mundane hyperparameter: the number of PPO epochs per batch. Increasing the number of epochs adds extra gradient steps on low-likelihood sequences after the high-likelihood ones saturate, amplifying training signal for unlikely solutions. Tuning this often-ignored hyperparameter is a complementary approach to the unlikeliness reward, and offers insight into the optimization dynamics that can lead to distribution sharpening.   

We demonstrate that our revised training recipe substantially improves pass@$N$ metrics across a range of values for $N$, while also substantially outperforming standard expert iteration. We combine unlikeliness reward and our insights into PPO epochs into a full recipe for reinforcement learning in formal theorem proving. We apply our recipe to theorem proving in Lean, resulting in a fully open pipeline that achieves competitive performance with DeepSeek-Prover-V1.5-RL on the miniF2F-test benchmark.

\section{Problem Setup}

We study the problem of training a language model for formal theorem proving, where the goal is to generate valid proofs of theorems in a proof assistant.
We use Lean~\citep{lean},
a proof assistant based on dependent type theory that supports the construction and verification of mathematical proofs. Lean has recently attracted interest in the AI and mathematics communities (e.g., \citet{yang2024formalmathematicalreasoningnew,tao2025MachineAssistedProof}).

Let \(\mathcal{D} = \{x_i\}_{i=1}^M\) be a dataset of theorem statements. Each statement consists of a natural language description and a formal statement expressing the theorem in Lean. Let \(R\) denote the verifier, which also functions as the reward function. Given a theorem statement \(x\) and a candidate proof \(y\), the Lean verifier returns a binary reward indicating whether \(y\) constitutes a successful proof of \(x\):
\[
    R(x, y) = \mathbbm{1}\{y \text{ proves } x\}.
\]
We assume access to an initial prover model \(\pi_{\text{base}}(y \mid x)\), a large language model (LLM) with some basic capability to generate proofs. Given a theorem statement \(x\), the model samples a completion \(y\) that attempts to prove the statement. 
Our goal is to fine-tune this model to improve its proof success rate, using problem instances from \(\mathcal{D}\) and the reward signal provided by \(R\).

\subsection{Evaluation Metric}
To evaluate the prover’s performance, we use the $\mathbf{pass@N}$ metric, which measures the probability that at least one of \(N\) independently sampled proof attempts succeeds. This metric is widely adopted in prior work due to its simplicity and close alignment with the practical use case of generating and verifying many proof attempts per theorem to find at least one that succeeds. 

Let \(x \in \mathcal{D}_{\text{test}}\) be a theorem, and let \(\{y_j\}_{j=1}^N \sim \pi_\theta(\cdot \mid x)\) denote \(N\) independent samples drawn from the model. The empirical \(\text{pass@}N\) metric for a single theorem is defined as:
\[
    \text{pass@}N(x; \pi_\theta) = \mathbbm{1}\left\{ \max_{1 \leq j \leq N} R(x, y_j) = 1 \right\}
\]
The average \(\text{pass@}N\) score on a test set \(\mathcal{D}_{\text{test}} = \{x_i\}_{i=1}^M\) is the average over individual theorems:
\[
    \text{pass@}N(\pi_\theta) = \frac{1}{M} \sum_{i=1}^M \text{pass@}N(x_i; \pi_\theta)
\]
In the context of reinforcement learning, a high \(\text{pass@}N\) also indicates that we are likely to receive a positive reward signal when sampling $N$ completions per problem.

\subsection{Reinforcement Learning}

We use Group Relative Policy Optimization (GRPO) as the foundation of our reinforcement learning experiments. GRPO was introduced by \cite{shao2024deepseekmathpushinglimitsmathematical} and has been successfully applied to train models such as DeepSeek-R1 and DeepSeek-Prover-V1.5-RL \citep{deepseekai2025deepseekr1incentivizingreasoningcapability, xin2024deepseekproverv15harnessingproofassistant}, showing strong performance in both informal and formal settings. 

GRPO is an extension of Proximal Policy Optimization (PPO) \citep{schulman2017proximalpolicyoptimizationalgorithms} that omits the critic model. For each question $x$, GRPO samples a group of outputs $\{y_1, \dots, y_G\} \sim \pi_{\theta_{old}}(y \mid x)$ from the current policy and maximizes the following objective:
\begin{align*}
&\mathcal{J}_{GRPO}(\theta) \\
& = \frac{1}{G} \sum_{i=1}^G \min \Bigg( \frac{\pi_\theta (y_i \mid x)}{\pi_{\theta_{old}}(y_i \mid x)} A_i, \\
& \quad\quad \text{clip} \left(\frac{\pi_\theta (y_i \mid x)}{\pi_{\theta_{old}}(y_i \mid x)}, 1-\epsilon, 1+\epsilon \right) A_i \Bigg) \\
& \quad - \beta_\mathrm{KL} \mathcal{D}_{\mathrm{KL}}[\pi_\theta \parallel \pi_{\text{ref}}]
\end{align*}

GRPO differs from PPO in how it computes the advantages $A_i$. Instead of subtracting a baseline predicted by the critic model, GRPO normalizes rewards within the group of samples. Let $r_i = R(x, y_i)$, then the advantages are computed as:
\[
    A_i = \frac{r_i - \text{mean}(\{r_1, \dots, r_G\})}{\text{std}(\{r_1, \dots, r_G\})}
\]

Note that when all or none of the samples solve the problem, $A_i = 0$ for all $i$ and there is no gradient with respect to model parameters $\theta$ (except for the KL term). To be more efficient with model updates, we implement a trick similar to Dynamic Sampling \citep{yu2025dapoopensourcellmreinforcement}. We maintain a buffer of recent samples that have nonzero advantage and only perform model updates once the buffer reaches the target batch size. 

\section{Does GRPO Improve Pass@N?}

We begin by investigating how GRPO behaves when applied to formal theorem proving. Our setup closely follows \citet{xin2024deepseekproverv15harnessingproofassistant} in terms of model choice and hyperparameter settings, though we curate our own dataset, as theirs has not been released. 

\subsection{Dataset}
The Lean Workbook dataset is a large-scale collection of approximately 140K Lean 4 theorem statements that were auto-formalized from natural language math problems \citep{ying2024leanworkbooklargescalelean}. Since unsolvable problems do not provide useful gradients during RL, we select a 10K subset of problems that were found to be solvable in \citet{wu2024internlm25stepproveradvancingautomatedtheorem}. These statements are still moderately challenging, as the solutions were discovered through an extremely compute-intensive search process. In addition, we also include the 244 problems from miniF2F-valid \citep{zheng2021minif2f}. 

From this combined dataset, we hold-out 200 theorems for validation, leaving 9.6K for training. Although miniF2F-test \citep{zheng2021minif2f} is a standard benchmark for theorem proving, we found high variance and inconsistent results on it when training at our scale, likely due to distribution shift and large difficulty gaps between problems. Thus, we primarily evaluate on our I.I.D. held-out set ($\mathcal{D}_{\text{val}}$) and only use miniF2F-test for our final large-scale experiments. We will refer to our training and validation sets as $\mathcal{D}_{\text{train}}$ and $\mathcal{D}_{\text{val}}$, respectively. 

\subsection{Training}
Our implementation of GRPO is built on the verl framework \citep{sheng2024hybridflow}, with modifications to support reward feedback from the Lean REPL. We use the Python wrapper for the Lean REPL released by \citet{xin2024deepseekproverv15harnessingproofassistant}, which we found to be more robust than previous open-source alternatives. The base model is DeepSeek-Prover-V1.5-SFT, which has moderate theorem-proving capabilities \citep{xin2024deepseekproverv15harnessingproofassistant}. We adopt the hyperparameters reported in \citet{xin2024deepseekproverv15harnessingproofassistant} where available: 
\begin{itemize}
    \item Learning rate = 5e-6
    \item KL loss coefficient = 0.02
    \item Number of samples per problem = 32.
\end{itemize}
However, we found the original learning rate to be unstable and use a reduced value of 1e-6. Due to compute constraints, we only train for one epoch on $\mathcal{D}_{\text{train}}$ and truncate the response length to 512 tokens, which suffices for over 99.5\% of samples. 

\begin{figure}[t]
\centering
\includegraphics[width=\columnwidth]{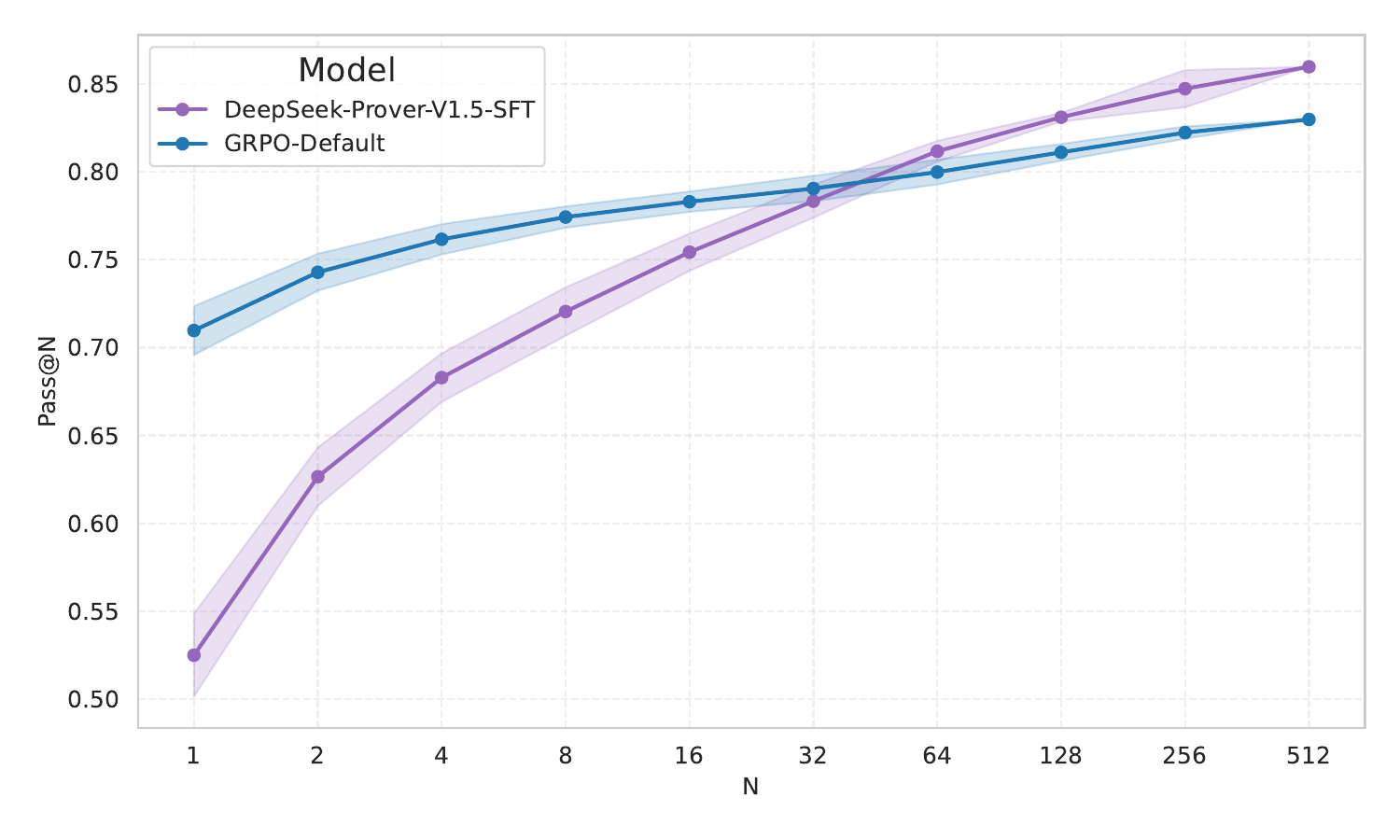}
\vspace{-2em}
\caption{Finetuning DeepSeek-Prover-V1.5-SFT with GRPO, evaluated on $\mathcal{D}_{\text{val}}$. GRPO improves pass@$N$ significantly for small $N$, but performs worse than the base model for large $N$. We aim to understand this behavior and develop methods to overcome it.}
\label{fig:initresults}
\vspace{-0.5em}
\end{figure}

\subsection{GRPO Fails to Improve Pass@N}
Figure~\ref{fig:initresults} presents model performance on $\mathcal{D}_{\text{val}}$, evaluated up to pass@512. GRPO substantially boosts pass@1 to pass@16, but the improvement diminishes for larger N. This pattern suggests that GRPO is effective at increasing the likelihood of already probable correct solutions but fails to surface new ones into the high-probability set, which is consistent with the findings of \citet{yue2025doesreinforcementlearningreally} and \citet{shao2024deepseekmathpushinglimitsmathematical}.  
Note that this is not an inherent failure of RL---boosting single-sample accuracy increases expected reward, but the benefit for formal theorem proving is limited. Next, we consider if and how RL can improve pass@$N$ at large $N$.

\subsection{Can RL Optimize Pass@N?}
In this section, 
we argue that improving pass@$N$ for large $N$ specifically requires RL to increase the probability of \textit{low-probability correct solutions} under the model.

Suppose that the initial model $\pi_0$ has a probability $p_{0}$ to solve a problem $x$, i.e.,
\[
   \sum_{y \text{ s.t. } R(x, y)=1} \pi_0(y \mid x) = p_{0}.
\]
The expected pass@N can then be expressed as:
\[
    \mathbb{E}[\text{pass@}N(\pi_0)] = 1 - (1 - p_0)^N.
\]
Now, we consider how RL training affects $p_0$. The exact outcome of taking gradient steps against the GRPO objective is impossible to predict analytically, but we can make estimates by assuming that we maximize the objective. 

For simplicity, we only consider early training steps, so that $\pi_{\theta_{old}} \approx \pi_0$, and disregard the KL term. The simplified GRPO objective is: 
\begin{align*} 
 &\mathcal{J}_{\mathrm{GRPO}}(\theta) \\
   & = \frac{1}{G}\sum_{i=1}^G \min \Bigg( \frac{\pi_\theta (y_i \mid x)}{\pi_{0}(y_i \mid x)} A_i,\\ & \text{clip}\left(\frac{\pi_\theta (y_i \mid x)}{\pi_{0}(y_i \mid x)}, 1-\epsilon, 1+\epsilon \right) A_i \Bigg). 
\end{align*}

\begin{figure}
\centering
\includegraphics[width=\columnwidth]{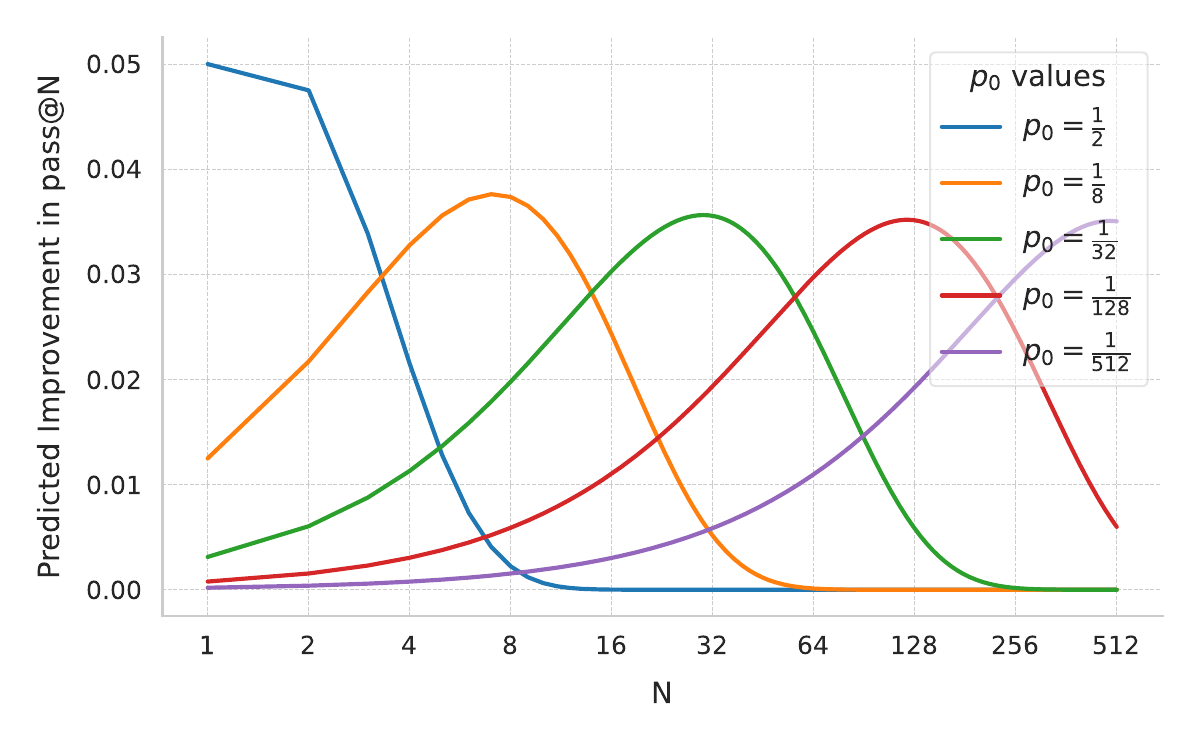}
\vspace{-2em}
\caption{Improvement in expected pass@$N$ assuming RL increases correct solution probabilities by a factor of $1 + \epsilon$ with $\epsilon=0.2$. Each curve corresponds to an initial $p_0 \in {1/2, 1/8, 1/32, 1/128, 1/512}$.}
\label{fig:passn}
\vspace{-0.5em}
\end{figure}

We make the simplifying assumption that the probability of each positive sample $y_+$ with $A_i > 0$ can be optimized independently. In the GRPO objective, each sample stops contributing gradient once $\pi_\theta(y_+ \mid x)/\pi_0(y_+ \mid x) \geq 1 + \epsilon$, thus we expect that the final ratio is close to the clipping bound:
\[\frac{\pi_\mathrm{RL} (y_+ \mid x)}{\pi_{0}(y_+ \mid x)} \approx 1 + \epsilon.\]
We can then predict the accuracy of the trained model:
\[p_\mathrm{RL} \approx (1 + \epsilon)p_0 \]
\[
\mathbb{E}[\text{pass@}N(\pi_\mathrm{RL})] \approx 1 - (1 - (1+\epsilon)p_0)^N.
\]
Figure~\ref{fig:passn} plots the expected improvement in pass@N for different initial $p_0$. When $p_0$ is large, the marginal gain in pass@512 is small. Conversely, when $p_0$ is small, gains are negligible for pass@1. In general, we see that increasing pass@$N$ requires the training algorithm to increase the probability of solutions with $p_0 \approx 1/N$. Thus, RL must specifically uplift the probability of \textit{low-probability correct solutions} to achieve improvements in pass@$N$ for large $N$. 

\subsection{Does GRPO Reinforce Unlikely Solutions?}
\label{sec:diagnose}
The analysis above, and our empirical observation that GRPO is not increasing pass@$N$, together suggest that GRPO may not be effectively uplifting low-probability correct solutions. To verify this, we examine training samples for the first 800 problems, computing their probabilities under the initial model and final GRPO-trained model.

Let $x_i$ be the i-th training problem and $y_{i, j}$ be the j-th corresponding solution. We compute $\pi_0(y_{i, j} \mid x_i)$ and $\pi_\mathrm{GRPO}(y_{i, j} \mid x_i)$ for all pairs. We are interested in whether $\pi_\mathrm{GRPO}(y_{i, j} \mid x_i)/\pi_0(y_{i, j} \mid x_i) \approx 1+\epsilon$, especially when $\pi_0(y_{i, j} \mid x_i)$ is small.

We find that the raw probability ratios are highly variable, containing extreme outliers, and the scale of $\pi_0(y_{i, j} \mid x_i)$ also differs widely across problems. This makes it difficult to analyze the raw model probabilities directly. Instead, we use the rank of a sample within its group as a proxy for its probability and consider the simpler, binary metric of whether $\pi_\mathrm{GRPO}(y_{i, j} \mid x_i)$ is greater than $ \pi_0(y_{i, j} \mid x_i)$. 

Formally, for each problem $x_i$, we sort the solutions $\{y_{i,1}, \dots, y_{i,G}\}$ in descending order of $\pi_0(y_{i,j} \mid x_i)$ to obtain $\{\tilde{y}_{i, 1}, \dots, \tilde{y}_{i, G}\}$. We are interested in the relationship between the rank of a solution and how likely it is to be uplifted by GRPO. For each rank $j \in \{1, \dots, G\}$, we compute the "uplift rate", averaging over positive samples:
\begin{align*}
u_j =\; &\mathop{\text{mean}}\limits_{i:\;R(x_i, \tilde{y}_{i,j})=1} \big( \\
        &\quad \mathbbm{1}\{\pi_\mathrm{GRPO}(\tilde{y}_{i, j} \mid x_i) > \pi_0(\tilde{y}_{i, j} \mid x_i)\} \big)
\end{align*}

\begin{figure}
\centering
\includegraphics[width=0.8\columnwidth]{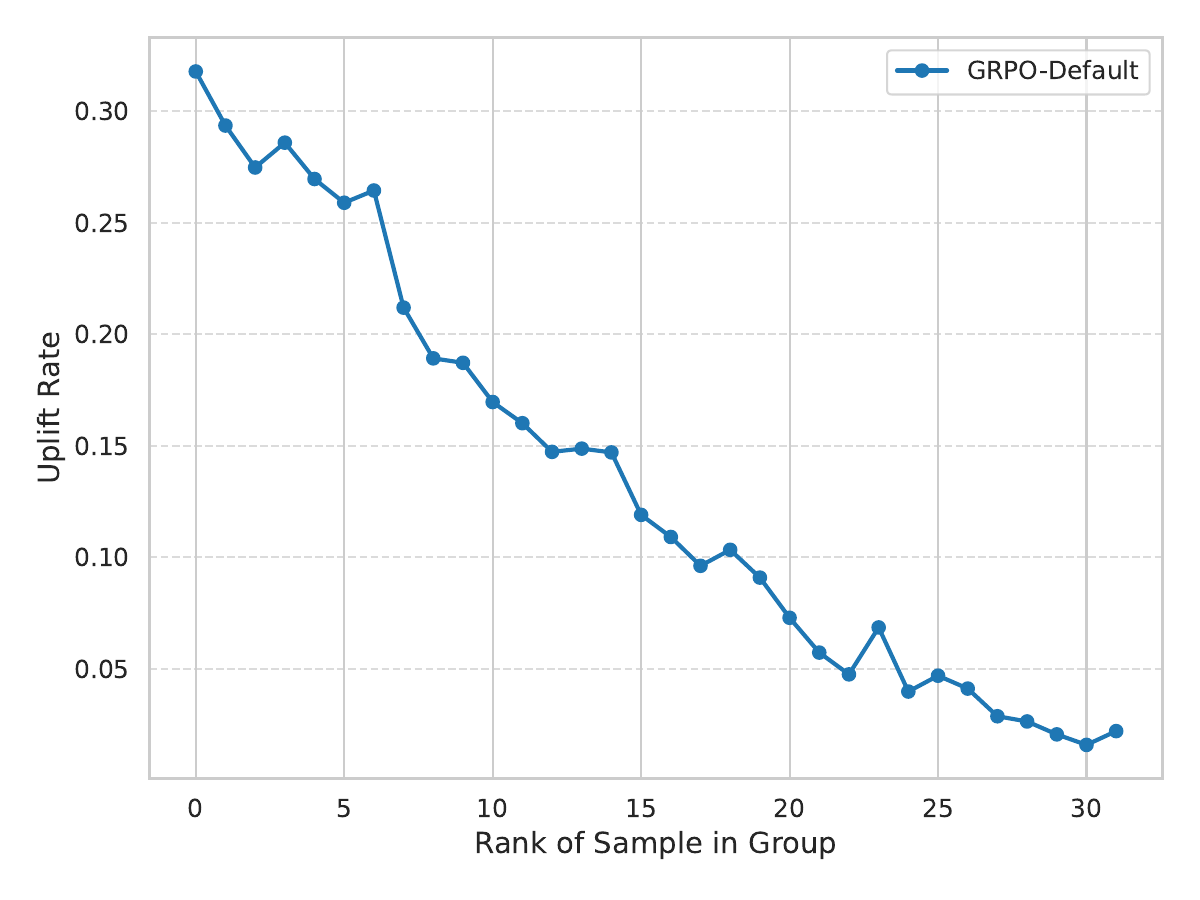}
\vspace{-0.5em}
\caption{Uplift rate $u_j$ as a function of rank $j$ among positive samples. GRPO rarely increases the probability of lowest-ranked (i.e. rarest) correct samples.}
\label{fig:rankbias}
\end{figure}

Figure~\ref{fig:rankbias} shows a clear positive correlation: GRPO is more likely to increase the probability of already high-probability correct solutions. In contrast, the low-probability positive samples -- those most critical for improving pass@$N$ at large $N$ -- are almost never uplifted. We confirm this behavior in a controlled toy environment (see Appendix~\ref{sec:toyenv}) and refer to this phenomenon as \textit{rank bias}.

\section{Improving GRPO for Multi-Sample Performance}
While the GRPO objective itself does not inherently favor high-probability solutions, our earlier analysis revealed a clear empirical bias: low-probability correct solutions are rarely reinforced. This behavior is counterintuitive -- when $\pi_0(y \mid x)$ is small, increasing the ratio $\pi_\mathrm{RL}(y \mid x)/\pi_0(y \mid x)$ requires less absolute probability mass and contributes equally to the GRPO objective. In principle, this should make low-probability solutions more attractive to optimize. The observed rank bias is therefore not a feature of the GRPO loss but likely a consequence of the optimizer’s biases.

In this section, we introduce the \textit{unlikeliness reward} to directly counteract this implicit bias, with the goal of improving pass@$N$ performance at large $N$. We also provide complementary analysis on the effect of certain hyperparameters on rank bias, which we later incorporate into our overall training recipe. 

\subsection{Unlikeliness Reward}
To explicitly correct for rank bias, we propose the \textbf{unlikeliness reward} -- a simple modification to the reward function that discourages reinforcing already high-probability solutions. For a group of samples $y_1, \dots, y_G$, let $\text{rank}(y_i) \in \{1, 2, \dots, G\}$ denote the rank of $y_i$ under the current policy $\pi_{\theta_{old}}(y_i \mid x)$, with rank 0 corresponding to the highest-probability sample. We modify the reward to be
\[
    r_i = R(x, y_i)\left(1 - \beta_{\text{rank}} \frac{G - \text{rank}(y_i)}{G} \right).
\]
A multiplicative penalty is applied to higher-probability solutions, increasing the relative advantage of rarer positive samples. Incorrect solutions remain unaffected, receiving $r_i=0$ regardless of rank. The coefficient $\beta_{\text{rank}}$ controls the strength of this perturbation; we fix $\beta_{\text{rank}} = 0.25$ in our experiments. 

Moreover, we continue to skip all samples that have zero advantage \textit{before} the perturbation. This ensures that no batch is dominated solely by the unlikeliness reward, and $R(x, y_i)$ still determines the direction of optimization for each sample. 

\subsection{Effects of PPO Epochs}
\label{sec:ppoepochs}
In addition to perturbing rewards, we find that increasing the number of optimization steps per sample (\textbf{ppo-epochs}) also mitigates rank bias. Standard implementations of PPO and GRPO typically use a single optimization step per batch 
 \citep{gpt_accelera, sheng2024hybridflow, yu2025dapoopensourcellmreinforcement}, which we found to produce biased updates. When taking multiple gradient steps, the initial steps may push high-rank solutions beyond the clipping threshold, so that subsequent steps are forced to focus on low-rank samples that are still unclipped. In this way, increasing ppo-epochs indirectly amplifies learning signal for low-rank samples. 
 
However, increasing ppo-epochs makes training substantially slower (Appendix~\ref{sec:traintime}) and potentially unstable. Thus, we prefer the unlikeliness reward as the more direct and efficient solution to address rank bias. 

\section{Experiments}
\label{sec:main_experiments}
For our main experiments, we use $\mathcal{D}_{\text{train}}$ and $\mathcal{D}_{\text{val}}$ for training and evaluation. We compare several GRPO variants with different hyperparameter settings, summarized in Table~\ref{tab:grpo_variants}. We increase the KL penalty because we found that it helps prevent deteriorating pass@$N$, but this change alone was not enough to improve pass@$N$ substantially (discussed in Appendix~\ref{appendix:kl}). All unlisted hyperparameters are kept the same.

\begin{table}[h]
  \centering
  \begin{tabular}{lccc}
    \hline
    \textbf{Model} & $K$ & $\beta_\mathrm{KL}$ & $\beta_\mathrm{rank}$ \\
    \hline
    GRPO-Default       & 1 & 0.02 & --    \\
    GRPO-Unlikeliness-1  & 1 & 0.10 & 0.25  \\
    GRPO-Unlikeliness-2  & 2 & 0.10 & 0.25  \\
    GRPO-Epochs-2      & 2 & 0.10 & --    \\
    GRPO-Epochs-3      & 3 & 0.10 & --    \\
    \hline
  \end{tabular}
  \caption{Hyperparameter settings for GRPO variants in our experiments. $K$ is the number of PPO epochs.}
  \label{tab:grpo_variants}
\end{table}

\begin{figure*}[t]
    \centering
    \includegraphics[width=0.7\textwidth]{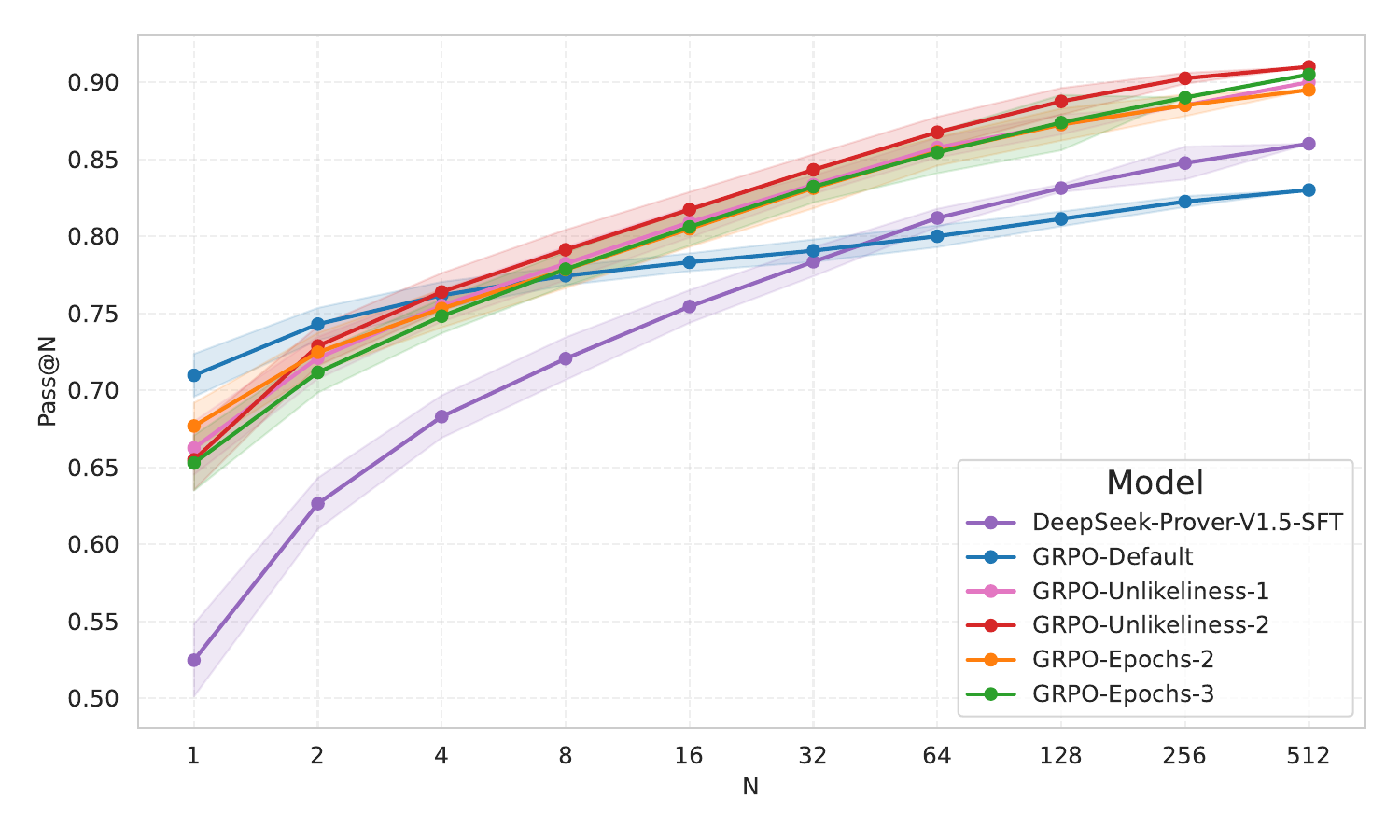}
    \vspace{-1.5em}
    \caption{Performance of GRPO variants on $\mathcal{D}_\text{val}$. Both the unlikeliness reward and additional PPO epochs improve pass@N. Appendix~\ref{appendix:eval} details how we compute these metrics.}
    \label{fig:mainresults}
\end{figure*}

\subsection{Results: Pass@N}
Figure~\ref{fig:mainresults} shows the performance of GRPO variants evaluated on $\mathcal{D}_\text{val}$. 
Introducing the unlikeliness reward 
leads to substantial improvements in pass@N at large N, with a minor tradeoff in pass@1 and pass@2. Interestingly, increasing PPO epochs also leads to improvements, consistent with our analysis in Section~\ref{sec:ppoepochs}. However, increasing PPO epochs leads to a significant increase in training time (Appendix~\ref{sec:traintime}). 

We also track the cumulative accuracy of the 32 samples generated per problem during training, including the baseline performance of a static model with no updates. Table~\ref{tab:cumulative_accuracy} reports the number of problems solved by each variant. All GRPO variants outperform the static model, with GRPO-Unlikeliness-2 solving the most problems. Since training runs for only one epoch, each example is effectively unseen at the time of sampling, indicating generalization within the epoch. 

\begin{table}
  \centering
  \begin{tabular}{lcc}
    \hline
    \textbf{Model} & \textbf{Solved} & \textbf{$\Delta$ Static} \\
    \hline
    Static (V1.5-SFT) & 7707 / 9600 & -- \\
    GRPO-Default                      & 7860 / 9600 & +153 \\
    GRPO-Epochs-2                    & 8008 / 9600 & +301 \\
    GRPO-Epochs-3                    & 8006 / 9600 & +299 \\
    GRPO-Unlikeliness-1                & 8023 / 9600 & +316 \\
    GRPO-Unlikeliness-2                & \textbf{8065 / 9600} & \textbf{+358} \\
    \hline
  \end{tabular}
  \caption{Number of training problems solved during one epoch on $\mathcal{D}_\text{train}$. GRPO variants improve over the static model, with GRPO-Unlikeliness-2 achieving the largest gain.}
  \label{tab:cumulative_accuracy}
\end{table}

\subsection{Analysis: Rank Bias}
\begin{figure}[t]
    \centering
    \includegraphics[width=\columnwidth]{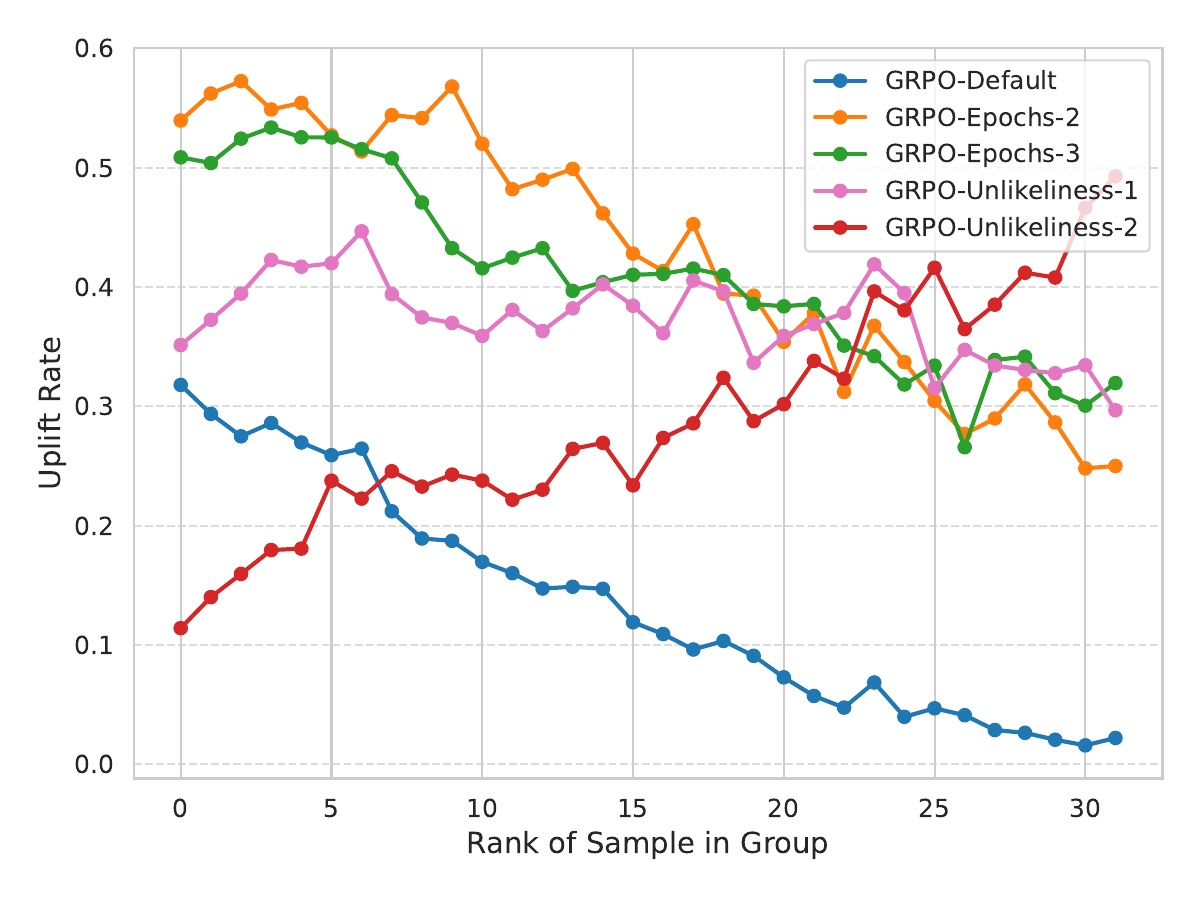}
    \vspace{-2em}
    \caption{Uplift rate $u_j$ as a function of rank $j$ for GRPO variants. The proposed methods improve the rate of reinforcing low-probability correct solutions.}
    \label{fig:uplift_new}
\end{figure}

To assess whether the proposed methods mitigate rank bias, we repeat the analysis from Section~\ref{sec:diagnose} by computing the $u_j$ metrics over the training samples for each GRPO variant. The results, shown in Figure~\ref{fig:uplift_new}, indicate substantial changes in GRPO’s behavior. GRPO-Unlikeliness-2 reverses the original pattern and is more likely to reinforce low-probability solutions. We also show that unlikeliness reward mitigates rank bias in our controlled environment (see Appendix~\ref{apx:ssec:unlikeliness-synthetic}).

In GRPO-Epochs-2 and GRPO-Epochs-3, the bias remains, but the overall strength of reinforcement is increased so that low-probability solutions are also sufficiently uplifted. 

\begin{figure}[t]
    \centering
    \includegraphics[width=\columnwidth]{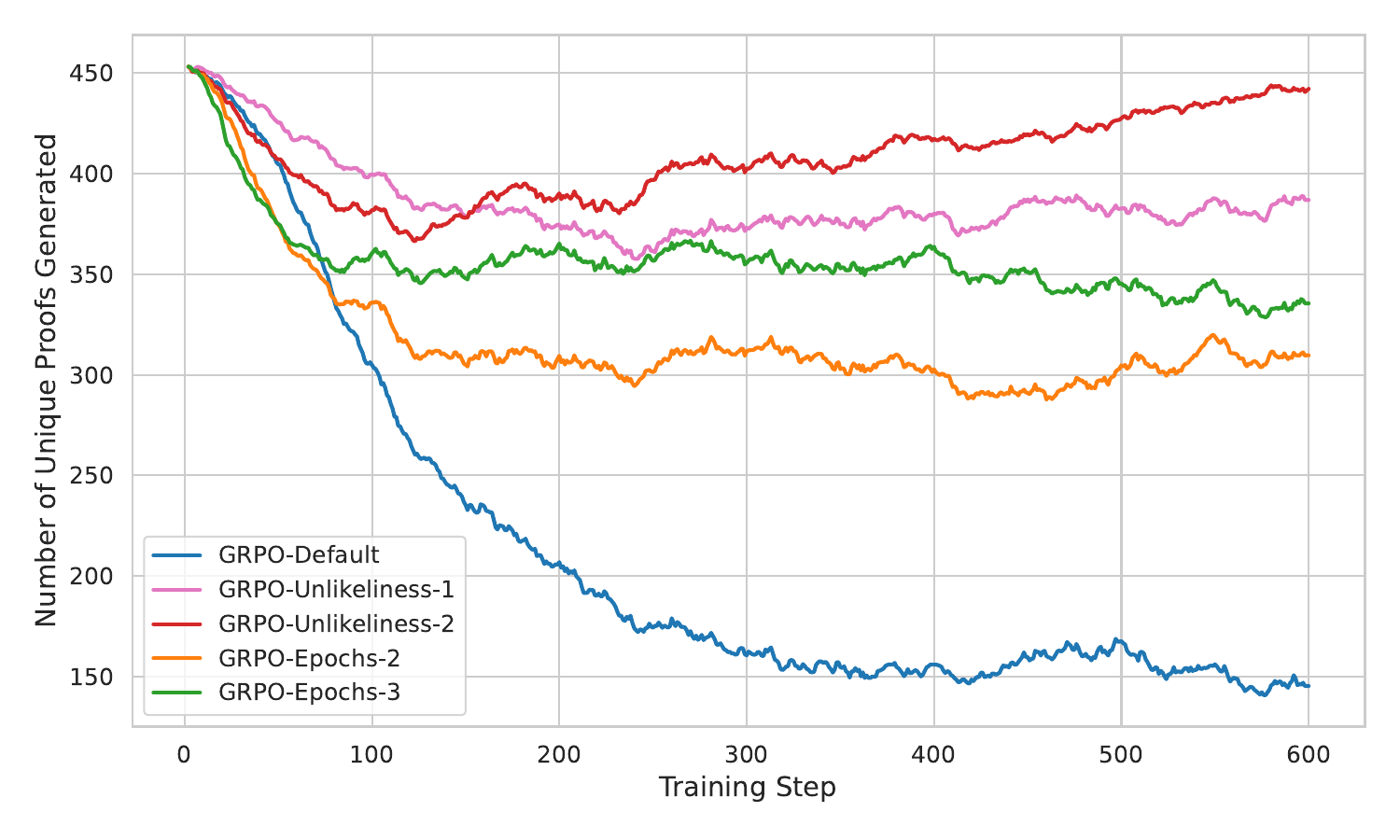}
    \vspace{-2em}
    \caption{Number of unique proofs generated at each training step (smoothed with EMA). Unlikeliness reward significantly improves sample diversity during training.}
    \label{fig:diversity}
\end{figure}

\subsection{Analysis: Sample Diversity}
Throughout training, we track the number of unique proofs generated per step, shown in Figure~\ref{fig:diversity}. GRPO-Unlikeliness-2 exhibits unique dynamics where diversity initially drops but later recovers, unlike other variants where diversity declines monotonically. This may reflect a self-correcting mechanism: initially dominant solutions are penalized, allowing low-probability correct solutions to resurface. This continuous rebalancing helps preserve a broad distribution of strategies throughout training.

We also observe that higher PPO epochs consistently increases sample diversity, up to ppo-epochs $=4$ where training becomes unstable. While this may seem counterintuitive -- since more optimization steps deviate the model further from its initial distribution -- it aligns with our earlier analysis. Higher PPO epochs indirectly amplifies rare solutions, thereby mitigating the sharpening effect typically caused by GRPO updates.

\subsection{Putting It All Together}
Finally, we evaluate \textbf{GRPO-Unlikeliness-2} in a large-scale experiment. We train the model on a dataset of 11k theorems, a larger and more challenging subset of Lean-Workbook that was solved and released by \citet{lin2025goedelproverfrontiermodelopensource}, making sure to exclude theorems in $\mathcal{D}_\text{val}$. 
We evaluate the resulting model on MiniF2F-test \citep{zheng2021minif2f}, a widely recognized benchmark for neural theorem proving, as well as $\mathcal{D}_\text{val}$. As reported in Table~\ref{tab:largescale}, \textbf{GRPO-Unlikeliness-2} achieves competitive results compared to DeepSeek-Prover-V1.5-RL \citep{xin2024deepseekproverv15harnessingproofassistant} on both datasets.

\begin{table}[t]
  \centering
  \begin{tabular}{lcc}
    \hline
    \textbf{Model} & \textbf{pass@32} & \textbf{pass@128} \\
    \hline
    \multicolumn{3}{c}{MiniF2F-test} \\
    V1.5-SFT & $47.1\pm0.6\%$ & $49.2\pm0.6\%$ \\
    V1.5-RL & $49.2\pm0.6\%$ & $51.2\pm0.3\%$ \\
    Ours     & $48.8\pm0.7\%$ & $50.6\pm0.5\%$ \\
    \hline
    \multicolumn{3}{c}{$\mathcal{D}_\text{val}$} \\
    V1.5-SFT & $78.3\pm0.9\%$ & $83.1\pm0.2\%$ \\
    V1.5-RL  & $84.8\pm0.9\%$ & $87.5\pm0.7\%$ \\
    Ours     &$84.3\pm0.9\%$ & $88.8\pm0.9\%$ \\
    \hline
  \end{tabular}
  \caption{pass@$N$ performance of our model compared to DeepSeek-Prover-V1.5-SFT and -RL from \citet{xin2024deepseekproverv15harnessingproofassistant} on MiniF2F-test and $\mathcal{D}_\text{val}$. Our model achieves competitive performance with DeepSeek-Prover-V1.5-RL while being fully open.}
  \label{tab:largescale}
\end{table}

\section{Related Work}
\textbf{Automated Theorem Proving:} 
\citet{polu2020generativelanguagemodelingautomated} pioneered transformer-based theorem provers that interact with proof assistants like Lean or Isabelle \citep{lean, isabella}. Subsequent work has developed state-tactic models \citep{polu2022formalmathematicsstatementcurriculum, wu2024internlm25stepproveradvancingautomatedtheorem, xin2025bfsproverscalablebestfirsttree} that generate one proof step at a time and full-proof models \citep{xin2024deepseekproverv15harnessingproofassistant, lin2025goedelproverfrontiermodelopensource} that produce complete proofs autoregressively, reducing interaction overhead.

Recent work has explored various directions in LLM-based theorem proving. \citet{lample2022hypertreeproofsearchneural}, \citet{xin2024deepseekproverv15harnessingproofassistant}, and \citet{xin2025bfsproverscalablebestfirsttree} explore the application of inference-time algorithms for proof discovery. 
\citet{jiang2023draftsketchproveguiding} and \citet{lin2025leanstarlearninginterleavethinking} use informal reasoning to guide formal proofs by integrating LLMs capable of reasoning in natural language. \citet{hu2024minictx} investigates training models that can incorporate novel context at test time. 
Our work is mainly focused on the post-training of theorem provers using reinforcement learning, which we detail next.

\textbf{Expert Iteration for Theorem Proving:}
Expert iteration alternates between search and learning \citep{anthony2017thinkingfastslowdeep}, and was first applied to theorem proving by \citet{polu2022formalmathematicsstatementcurriculum}. It has since become the dominant paradigm, appearing in recent work like \citet{wu2024internlm25stepproveradvancingautomatedtheorem}, \citet{xin2025bfsproverscalablebestfirsttree}, and \citet{lin2025goedelproverfrontiermodelopensource}. 
\citet{xin2025bfsproverscalablebestfirsttree} explores the viability of best-first search for data collection, while \citet{wu2024internlm25stepproveradvancingautomatedtheorem} and \citet{lin2025goedelproverfrontiermodelopensource} achieve state-of-the-art performance at the time by performing large-scale expert iteration on autoformalized theorem statements.

\textbf{RL for Theorem Proving:}
Compared to expert iteration, the use of more general RL algorithms is relatively underexplored.
A notable exception is \citet{xin2024deepseekproverv15harnessingproofassistant}, which showed GRPO can enhance a SFT model using only additional theorem statements and the verifier reward. In the low-data setting, \citet{gloeckle2024abel} successfully trained a strong theorem prover by adapting the AlphaZero algorithm~\cite{silver2017masteringchessshogiselfplay} to proof trees. \citet{xin2025bfsproverscalablebestfirsttree} used direct preference optimization~\citep{rafailov2023direct} in their pipeline, but only for the minor role of training against proof steps that cause immediate errors. 

More recent work has begun adapting techniques from OpenAI o1 \cite{openai2024openaio1card} and DeepSeek-R1 \cite{deepseekai2025deepseekr1incentivizingreasoningcapability} to train reasoning models for theorem proving \citep{wang2025kiminaproverpreviewlargeformal, ren2025deepseekproverv2advancingformalmathematical, zhang2025leanabellproverposttrainingscalingformal}. These works have achieved state-of-the-art performance by building models that can engage in long chain-of-thought style reasoning, either calling formal proof models as subroutines \citep{ren2025deepseekproverv2advancingformalmathematical} or devising hierarchical strategies to break down the problem \citep{wang2025kiminaproverpreviewlargeformal}.

\textbf{RL for Multi-Sample Performance:}
Several existing works specifically investigate the issue of RL's pass@$N$ performance.
\citet{yue2025doesreinforcementlearningreally} argues that instead of learning novel capabilities, RL with verifier reward mainly concentrates the model's outputs around correct answers already present in the base model's samples. Their experiments also show an improvement in pass@ small $N$ and deterioration at large $N$. 
\citet{chow2024inferenceawarefinetuningbestofnsampling} and \citet{tang2025optimizinglanguagemodelsinference} consider novel RL formulations that explicitly optimize for best-of-$N$ performance. They derive BoN-aware RL algorithms and demonstrate improved performance, but still consider a smaller range of $N$ (pass@32) than is typical in formal theorem proving. In the expert iteration setting, \citet{dang2025weightensemblingimprovesreasoning} identifies that pass@$N$ deteriorates due to diversity collapse and shows that interpolating model weights with an early checkpoint mitigates this issue.

Compared to these previous works, we are the first to attribute RL's poor multi-sample performance to an inability to reinforce low-probability samples. We also provide a simple and direct solution to address this issue and improve pass@$N$ performance.

\section{Conclusion}
We investigated GRPO's poor multi-sample performance in the setting of formal theorem proving, theorizing a connection between degraded pass@$N$ at large $N$ and the failure to reinforce low-probability solutions. Our analysis revealed an implicit bias in GRPO: it preferentially reinforces already high-probability sequences while largely ignoring rare but correct ones. To address this, we introduced the \textit{unlikeliness reward}, a simple yet effective modification that directly shifts reinforcement toward rare samples. 
Our experiments confirm that the unlikeliness reward enables GRPO to make significant gains in pass@$N$ at large $N$ and drastically improves sample diversity compared to existing methods. Using our revised recipe, we train a model that is competitive with DeepSeek-Prover-V1.5-RL and release our implementation publicly.

\section*{Limitations}
While we offer a lightweight solution for improving GRPO's multi-sample performance, future work could explore other strategies for uniformly reinforcing correct samples or for directly optimizing performance under specific inference-time algorithms. In particular, developing inference-aware reinforcement learning algorithms that are efficient to train remains an open direction. 

Moreover, recent applications of RL have shifted toward the reasoning paradigm, where models generate long reasoning paths often involving behaviors such as planning, backtracking, and self-critique. In these settings, the behavior of algorithms like GRPO may differ qualitatively due to the increased diversity and complexity of possible reasoning paths. We leave as future work to determine whether methods that amplify rare but correct solutions can similarly enhance exploration and generalization in reasoning models. 

\section*{Acknowledgements}
Sean Welleck thanks Convergent Research and the Lean FRO for their support. 

\bibliography{custom}

\appendix
\section{Toy Environment}
\label{sec:toyenv}
After observing that GRPO failed to improve pass@$N$ metrics, we constructed a simplified toy environment to isolate the issue and efficiently test potential solutions. This appendix details the design of the environment and presents our experimental results within it.

\subsection{Environment Design}
We design a minimalistic toy environment for rapid experimentation. The environment is fully observable, with state space $\mathcal{S} = \mathbb{R}^{10}$ and discrete action space $\mathcal{A} = \{1, \dots, 128\}$. Each action $a \in \mathcal{A}$ is associated with a fixed, randomly initialized but hidden vector $v_a \in \mathbb{R}^{10}$.

The binary reward function $R_\tau: \mathcal{S} \times \mathcal{A} \rightarrow \{0, 1\}$ is defined as:

$$
    R_\tau(s, a) = \mathbbm{1}\{s^\top v_a \geq \tau\}
$$

Here, $\tau$ is a threshold controlling environment difficulty. Higher $\tau$ values restrict the reward to fewer actions, thus increasing difficulty. We fix $\tau = 1.0$ during training but vary $\tau$ during evaluation to simulate different difficulty levels.

\subsection{Policy Model}
The policy model $\pi_\theta(a \mid s)$ is a simple two-layer multilayer perceptron (MLP) mapping state $s$ to a probability distribution over actions in $\mathcal{A}$.

\subsection{GRPO Training and Diagnosis}
We train the model using GRPO for 200 steps and evaluate pass@$N$ metrics at $N \in \{1, 4, 8, 16, 32\}$. Initial evaluations at training difficulty $\tau=1.0$ suggest GRPO improves pass rates across all $N$:
\begin{center}
\includegraphics[width=0.8\columnwidth]{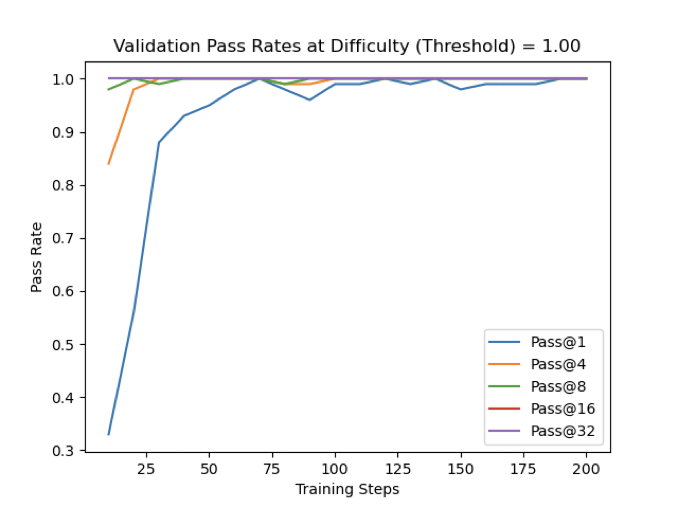}
\end{center}

However, evaluations at increased difficulties ($\tau=4.0$ and $\tau=5.0$) reveal pass@32 deteriorates over training, aligning with observations in the original setting:
\begin{center}
\includegraphics[width=0.8\columnwidth]{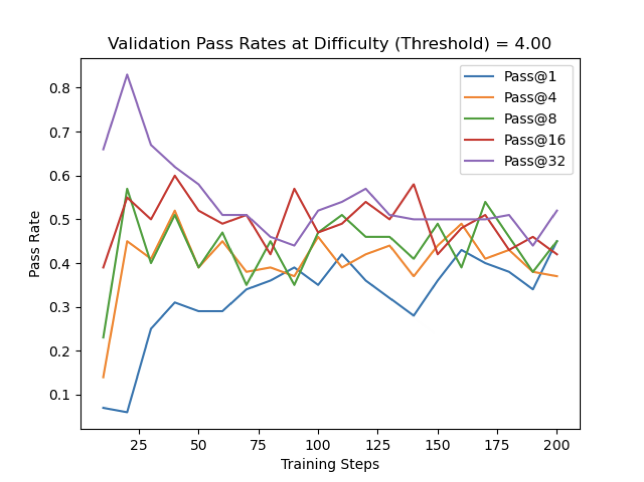}
\includegraphics[width=0.8\columnwidth]{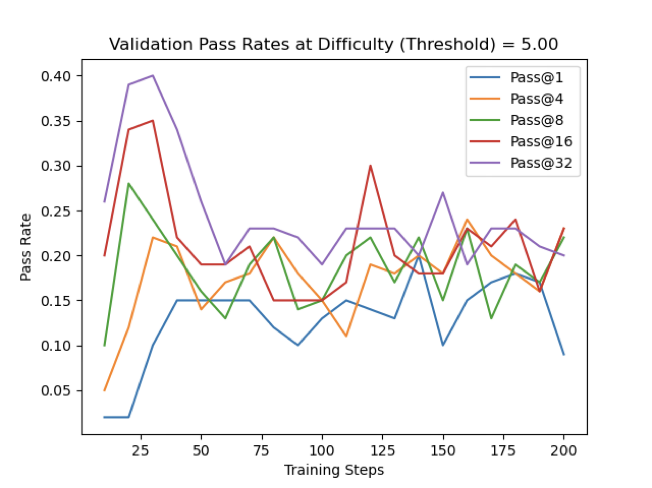}
\end{center}

Analyzing uplift rate metrics (Section~\ref{sec:diagnose}), we identify a rank bias in GRPO, showing preferential reinforcement of already high-probability solutions:
\begin{center}
\includegraphics[width=0.8\columnwidth]{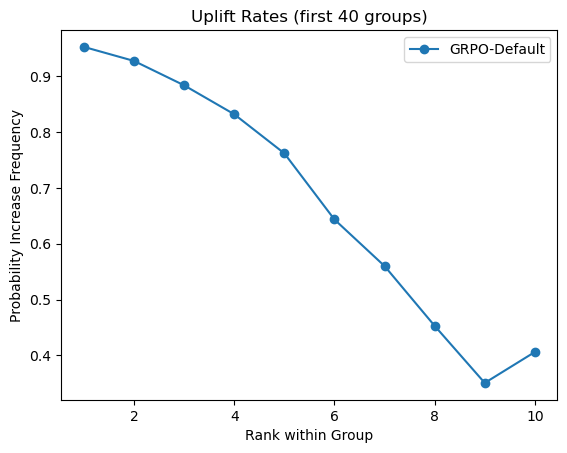}
\end{center}

\subsection{Unlikeliness Reward}
\label{apx:ssec:unlikeliness-synthetic}
We investigate the impact of unlikeliness reward within this toy environment. It effectively neutralizes the rank bias, making the uplift rates notably more uniform:
\begin{center}
\includegraphics[width=0.8\columnwidth]{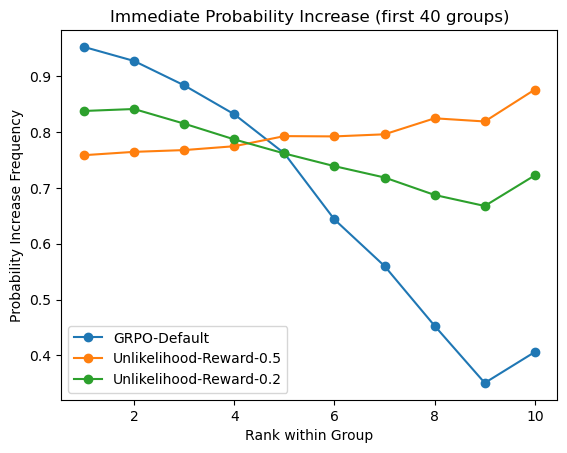}
\end{center}

Consequently, the unlikeliness reward significantly improves pass@32 performance in the difficult setting $\tau=5.0$, contrasting sharply with default GRPO, whose pass@32 performance declines to near chance levels:
\begin{center}
\includegraphics[width=0.8\columnwidth]{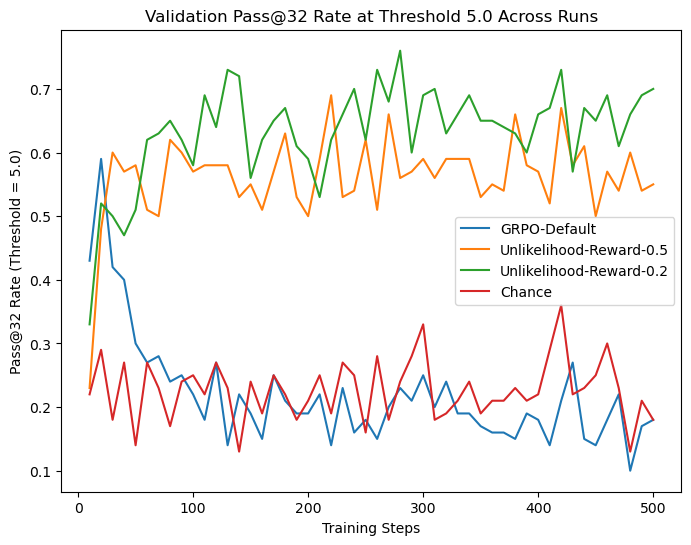}
\end{center}

Additionally, incorporating the unlikeliness reward substantially increases the entropy of the predicted action distribution:
\begin{center}
\includegraphics[width=0.8\columnwidth]{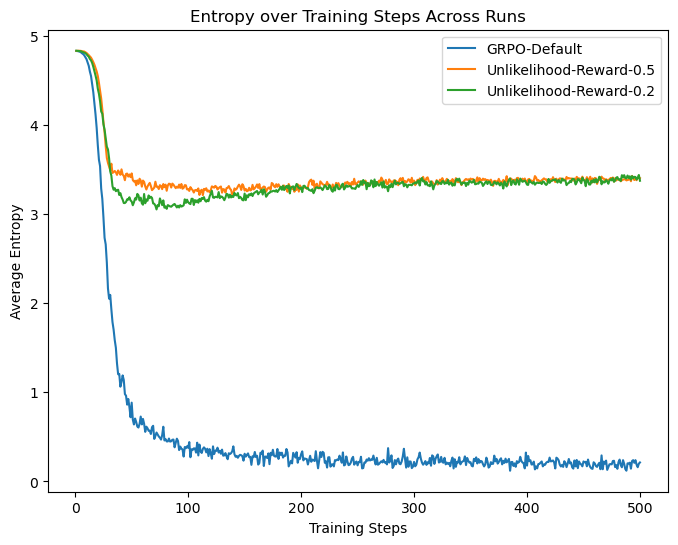}
\end{center}

\section{Training Setup}

The main experiments in Section~\ref{sec:main_experiments} are conducted on 4 NVIDIA L40S GPUs, with 500GB of RAM and 48–64 CPUs allocated for running parallel instances of the Lean REPL.

\subsection{Training Time}
\label{sec:traintime}

All training runs in the main experiment complete within 36 hours. Each training step primarily consists of three stages: sequence generation, proof verification, and policy model updates. The generation and verification stages are shared across all methods and take approximately 120 seconds per batch (16 problems × 32 attempts). The duration of the policy update step depends on the number of PPO epochs, as shown below:
\vspace{1em}
\begin{center}
  \begin{tabular}{lc}
    \hline
    \textbf{PPO Epochs} & \textbf{Policy Update Time (s)} \\
    \hline
    1 & $\approx$ 70 \\
    2 & $\approx$ 140 \\
    3 & $\approx$ 210 \\
    \hline
  \end{tabular}
\end{center}

\section{Evaluation Metrics}
\label{appendix:eval}
We begin by selecting a maximum sample size $N_\text{max}$ (512 in our experiments) and generate $N_\text{max}$ responses for each problem. To compute pass@\textit{n}, we divide the responses for each problem into $N_\text{max} / n$ chunks and assign each chunk a binary reward indicating whether any proof within it is valid. The $i$-th trial of pass@\textit{n} is then computed by averaging the binary rewards across the $i$-th chunk of all problems. We report the mean and standard deviation across trials. Note that for pass@512, there is only a single trial, so we omit the standard deviation in our plots.

\section{Effects of KL Penalty}
\label{appendix:kl}
Recent results have shown that the pass rates of theorem prover models can continue to improve with increased sampling, up to hundreds of thousands of passes \citep{lin2025goedelproverfrontiermodelopensource}. This suggests that the distribution of the base model is highly diverse and crucial to preserve during fine-tuning. Prior work addressed this in the SFT setting by ensembling fine-tuned model weights with the original \citep{dang2025weightensemblingimprovesreasoning}. Since GRPO already has a regularization mechanism through the KL penalty, we simply increase the KL loss coefficient to $0.1$ to better preserve the original distribution.

To isolate the contribution from unlikeliness reward and PPO epochs, we conduct a control run that only increases the KL penalty from GRPO-Default. This corresponds to an additional row for Table~\ref{tab:grpo_variants}:

\vspace{1em}
  \begin{tabular}{lccc}
    \hline
    \textbf{Model} & $K$ & $\beta_\mathrm{KL}$ & $\beta_\mathrm{rank}$ \\
    \hline
    GRPO-High-KL  & 1 & 0.10 & --    \\
    \hline
  \end{tabular}
\vspace{1em}


\begin{figure}
    \centering
    \includegraphics[width=\columnwidth]{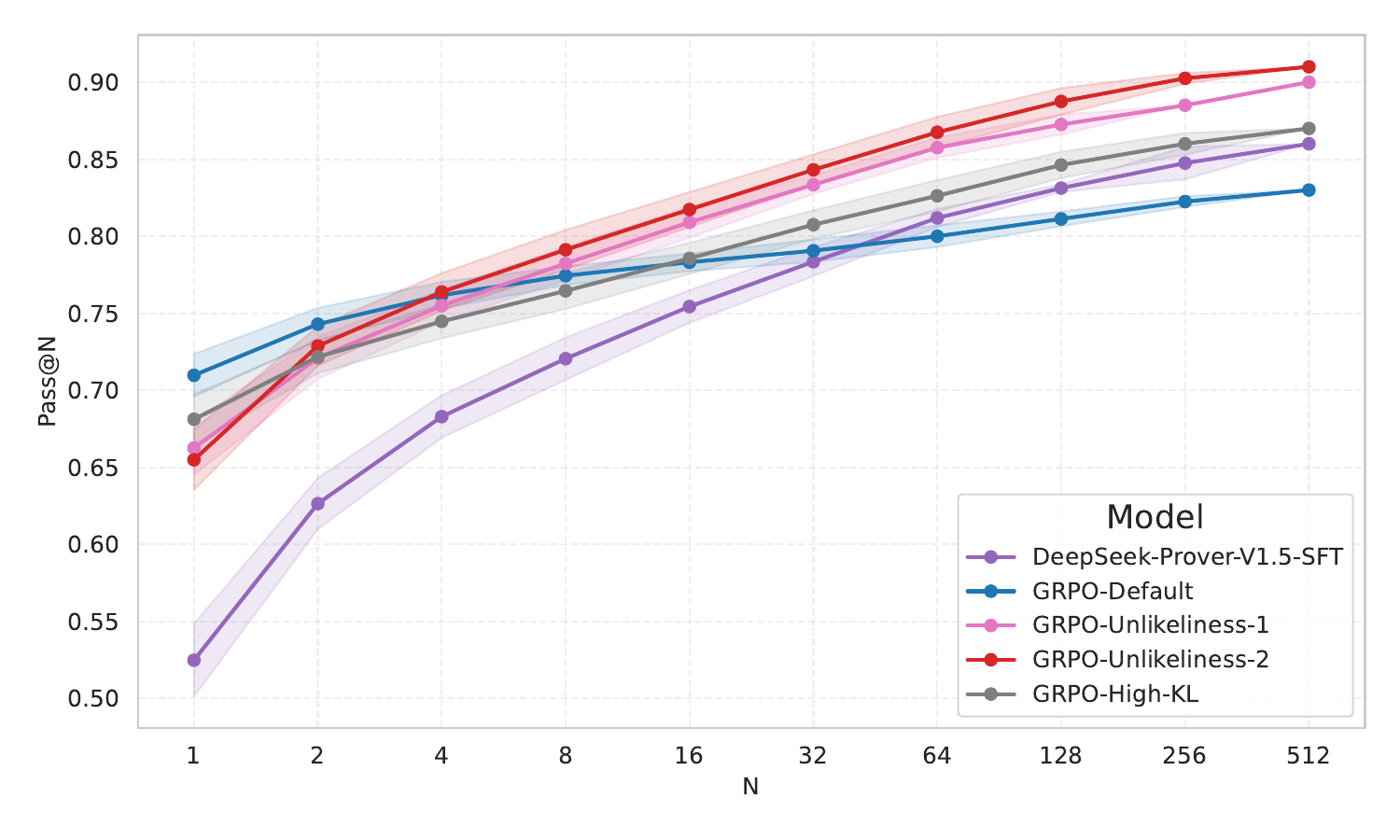}
    \vspace{-2em}
    \caption{Performance of GRPO variants including GRPO-High-KL on $\mathcal{D}_\text{val}$. For readability, we omit some variants.}
    \label{fig:passkl}
\end{figure}

\begin{figure}
    \centering
    \includegraphics[width=0.8\columnwidth]{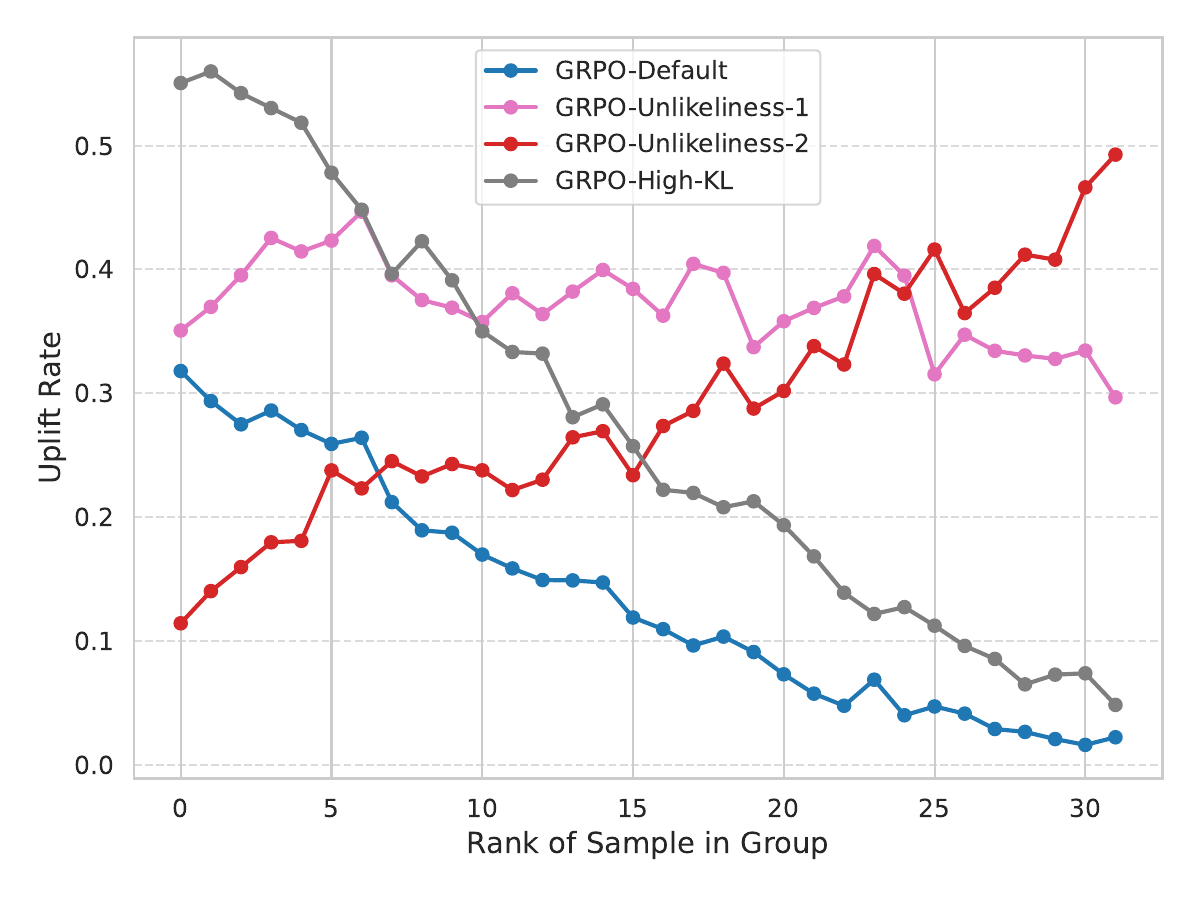}
    \vspace{-1em}
    \caption{Uplift rates of GRPO variants including GRPO-High-KL.}
    \label{fig:upliftkl}
\end{figure}

We find that, while this change prevented the deterioration of pass@$N$ performance, it did not bring a substantial improvement over the base model (Figure~\ref{fig:passkl}). This is likely because the RL updates still fail to uplift low-rank samples (Figure~\ref{fig:upliftkl}). Thus, we treat KL regularization as a supporting modification rather than a solution in itself.

\end{document}